\documentclass{bmvc2k}
\usepackage{amsfonts}
\usepackage{amsmath}
\usepackage{amssymb}
\usepackage{array}
\usepackage{booktabs}
\usepackage{dirtytalk}
\usepackage{mathtools}
\usepackage{microtype}
\usepackage{multirow}
\usepackage{nicefrac}
\usepackage{url}
\usepackage{amsfonts}
\usepackage{xcolor}
\usepackage{rotating}
\usepackage{pifont}

\newcommand{\cmark}{\ding{52}}%
\newcommand{\xmark}{\ding{56}}%

\title{Dual Graph Convolutional Network for Semantic Segmentation}

\addauthor{Li Zhang*}{lz@robots.ox.ac.uk}{1}
\addauthor{Xiangtai Li*}{lxtpku@pku.edu.cn}{2}
\addauthor{Anurag Arnab}{aarnab@robots.ox.ac.uk}{1}
\addauthor{Kuiyuan Yang}{kuiyuanyang@deepmotion.ai}{3}
\addauthor{Yunhai Tong}{yhtong@pku.edu.cn}{2}
\addauthor{Philip H.S. Torr}{phst@robots.ox.ac.uk}{1}
\addinstitution{
Department of Engineering Science,\\
Torr Vision Group,\\
University of Oxford
}
\addinstitution{
Key Laboratory of Machine Perception, \\
School of EECS,\\
Peking University
}
\addinstitution{
DeepMotion AI Research
}

\runninghead{Zhang \etal}{Dual Graph Convolutional Network}

\def\eg{\emph{e.g}\bmvaOneDot}

\def\etal{\emph{et al}\bmvaOneDot}
\def\ie{\textit{i.e.}}

\begin{document}

\maketitle

\begin{abstract}
\noindent
Exploiting long-range contextual information is key for pixel-wise prediction tasks such as semantic segmentation.
In contrast to previous work that uses multi-scale feature fusion or dilated convolutions, we propose a novel graph-convolutional network (GCN) to address this problem.
Our \emph{Dual Graph Convolutional Network} (DGCNet) models the global context of the input feature by modelling two orthogonal graphs in a single framework.
The first component models spatial relationships between pixels in the image, whilst the second models interdependencies along the channel dimensions of the network's feature map.
This is done efficiently by projecting the feature into a new, lower-dimensional space where all pairwise interactions can be modelled, before reprojecting into the original space.
Our simple method provides substantial benefits over a strong baseline and achieves state-of-the-art results on both Cityscapes (82.0\% mean IoU) and Pascal Context (53.7\% mean IoU) datasets.
Code and models are made available to foster any further research (\url{https://github.com/lxtGH/GALD-DGCNet}).
\end{abstract}

\section{Introduction}

Semantic segmentation is a fundamental problem in computer vision, and aims to assign an object class label to each pixel in an image.
It has numerous applications including autonomous driving, augmented- and virtual reality and medical diagnosis.

An inherent challenge in semantic segmentation is that 
pixels are difficult to classify when considered in isolation, 
as local image evidence is ambiguous and noisy.
Therefore, segmentation systems must be able to effectively capture contextual information in order to reason about occlusions, small objects and model object co-occurrences in a scene.

Current state-of-the-art methods are all based on deep learning using fully convolutional networks (FCNs) \cite{fcn}.
However, the receptive field of an FCN grows slowly (only linearly) with increasing depth in the network, and its limited receptive field is not able to capture longer-range relationships between pixels in an image.
Dilated convolutions \cite{deeplabv1, dilation} have been proposed to remedy this.
However, the resulting feature representation is dominated by large objects in the image, and consequently, performance on small objects is poor.
Another direction has been to fuse multiscale features within the network \cite{pspnet,parsenet,hou2020strip} or to use LSTMs to propagate information spatially \cite{lstmseg,dagseg}.
Recently, several methods based on self-attention \cite{Nonlocal,DAnet,ocnet,li2019global} have also been used to learn an affinity map at each spatial position that propagates information to its neighbours.
However, the memory requirements of the large affinity matrix renders these methods unsuitable for high resolution imagery (such as the Cityscapes dataset \cite{Cityscapes}).

\begin{figure*}
	\centering
	\includegraphics[width=1.0\linewidth]{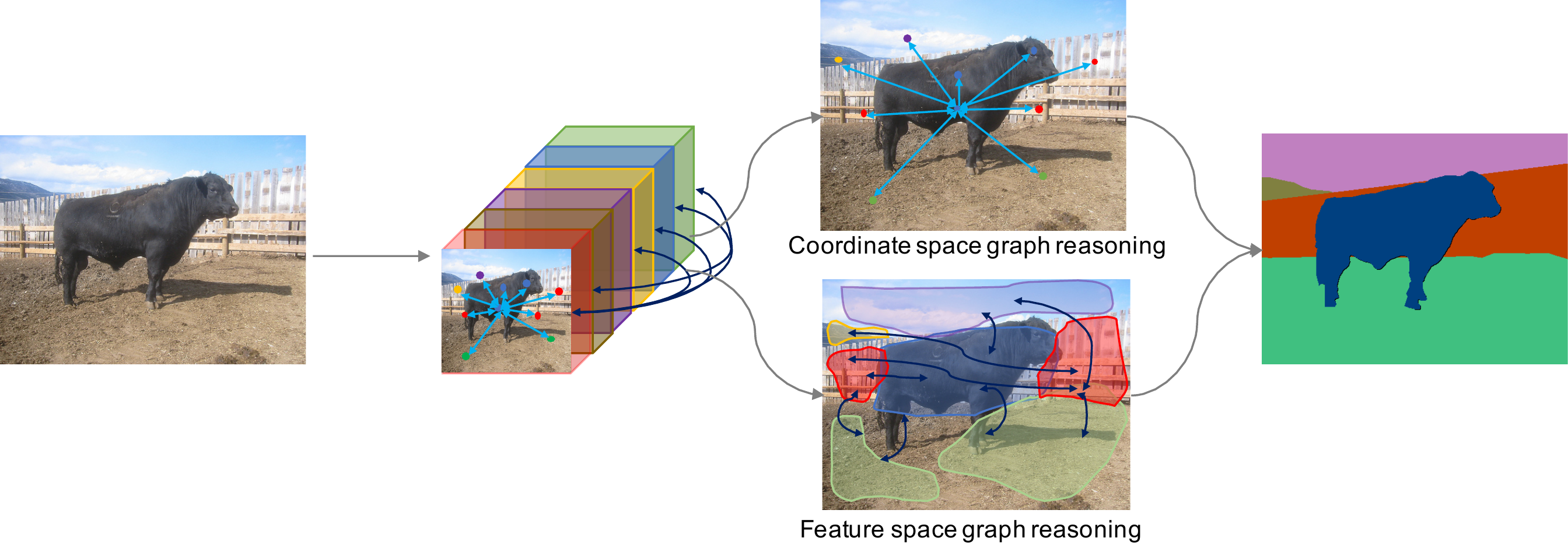}
	\caption{Our proposed \emph{DGCNet} exploits contextual information across the whole image by
	proposing a graph convolutional network to efficiently propagate information along both the spatial and channel dimensions of a convolutional feature map.}
	\label{fig:teaser}
\end{figure*}

In this paper, we use graph-convolutional networks (GCNs) \cite{gcnpaper} to effectively and efficiently model contextual information for semantic segmentation.
GCNs have recently been applied to scene understanding tasks \cite{beyond_grids,graph_reason,SGR_gcn,zhang2020dynamic}, as they are able to globally propagate information through the whole image in a manner that is conditional on the input.
This provides greater representational power than methods based on Conditional Random Fields \cite{zheng2015conditional, deeplabv1, arnab2018conditional} which have historically been employed for semantic segmentation \cite{he2004multiscale,shotton2006textonboost}.

As shown in Fig.~\ref{fig:teaser}, our proposed method consists of two primary components: the coordinate space GCN explicitly models the spatial relationships between pixels in the image, enabling our network to produce coherent predictions that consider all objects in the image, whilst the feature space GCN models interdependencies along the channel dimensions of the network's feature map. 
Assuming that filters in later layers of a CNN are responsive to object parts and other high-level features \cite{zeiler2014visualizing}, the feature space GCN captures correlations between more abstract features in the image like object parts. 
After reasoning, these two complementary relation-aware feature are distributed back to the original coordinate space and added to the original feature. 

Using our proposed approach, we establish new state-of-the-arts on the Cityscapes \cite{Cityscapes} (82.0\% mean IoU) and Pascal Context \cite{mottaghi_cvpr14} (53.7\% mean IoU) datasets.
\section{Related Work}

Following the success of deep neural networks for image classification \cite{alexnet,vgg,resnet},
recent works in semantic segmentation all leverage fully-convolutional networks (FCNs) \cite{fcn}.
A limitation of standard FCNs is their small receptive field which prevents them from taking all the contextual information in the scene into account.
The DeepLab series of papers \cite{deeplabv1, deeplabv2, deeplabv3} proposed atrous or dilated convolutions and atrous spatial pyramid pooling (ASPP) to increase the effective receptive field.
DenseASPP improved on \cite{deeplabv2} by densely connecting convolutional layers with different dilation rates.
PSPNet \cite{pspnet}, on the other hand, used a pyramid pooling module to fuse convolutional features from multiple scales.
Similarly, encoder-decoder network structures \cite{deconvnet,unet,fpn} combine features from early- and late-stages in the network to fuse mid-level and high-level semantic features.
Deeplab-v3+ \cite{deeplabv3p} also followed this approach by fusing lower-level features into its decoder.
Di \etal \cite{msci} also recursively, locally fused feature maps of every two levels in a feature pyramid into one.

Another approach has been proposed to more explicitly account for the relations between all pixels in the image.
DAG-RNN \cite{dagseg} models a directed acyclic graph with a recurrent network that propagates information.
PSANet \cite{psanet} captures pixel-to-pixel relations using an attention module that takes the relative location of each pixel into account.
On the other hand, EncNet \cite{encodingnet} and DFN \cite{dfn} use attention along the channel dimension of the convolutional feature map to account for global context such as the co-occurrences of different classes in the scene.

Following on from these approaches, graph neural networks \cite{gcnpaper} have also been used to model long-range context in the scene.
Non-local networks \cite{Nonlocal} applied this to video understanding and object detection by learning an affinity map between all pixels in the image or video frames.
This allowed the network to effectively increase its receptive field to the whole image.
The non-local operator has been applied to segmentation by OCNet \cite{ocnet} and DANet \cite{DAnet} recently.
However, these methods have a (sometimes prohibitively) high memory cost as the affinity matrix grows quadratically with the number of pixels in the image.
To bypass this problem, several works~\cite{graph_reason,SGR_gcn,beyond_grids} have modelled the dependencies between regions of the image rather than individual pixels.
This is done by aggregating features from the original ``co-ordinate space'' to a lower-dimensional intermediate representation, performing graph convolutions in this space, and then reprojecting back onto the original co-ordinate space.

However, differently from these recent GCN methods, we propose the Dual Graph Convolutional Network (DGCNet) to model the global context of the input feature by considering \emph{two orthogonal graphs in a single general framework}.
Specifically, with different mapping strategies, we first project the feature into a new coordinate space and a non-coordinate (feature) space where global relational reasoning can be computed efficiently.
After reasoning, two complementary relation-aware features are distributed back to the original coordinate space and added to the original feature. The refined feature thus contains rich global contextual information and can be further provided to the following layers to learn better task-specific representations.

\section{Methodology}
In this section, we first revisit the graph convolutional network in Sec.~\ref{sec:method_prelim} and then introduce the formulation of our proposed DGCNet in Sec.~\ref{sec:method_coordinate_space} and ~\ref{sec:method_feature_space}.
Finally, we detail the resulting network architecture in Sec.~\ref{sec:network_arch}.

\subsection{Preliminaries}
\label{sec:method_prelim}
\begin{figure*}
\vspace{5mm}
	\centering
	\includegraphics[width=0.95\linewidth]{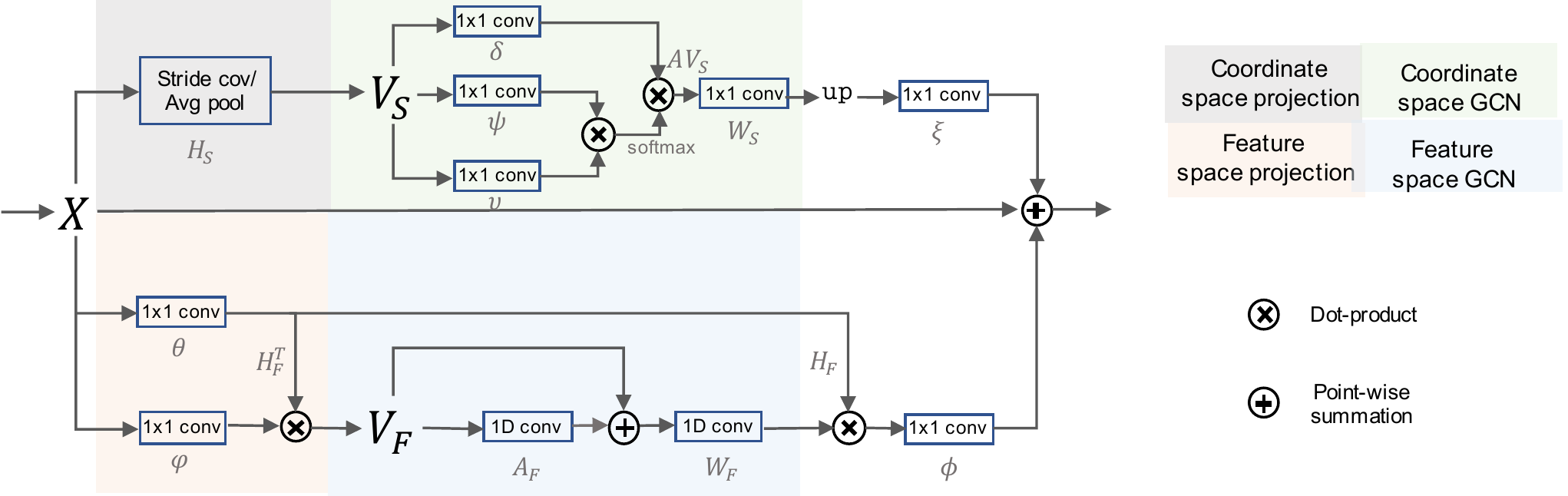}
	\vspace{-2mm}
	\caption{Illustration of our proposed DGCNet.
	        Our method consists of two branches, which each consist of a graph convolutional network (GCN) to model contextual information in the spatial- and channel-dimensions in a convolutional feature map, $X$.
	}
	\label{fig:net}
\end{figure*}

\noindent{\textbf{Revisiting the graph convolution.}} 
Assume an input tensor $\mathbf{X} \in \mathbb{R}^{N \times D}$, where $D$ is the feature dimension and $N = H \times W$ is the number of locations defined on regular grid coordinates $\Omega=\{1,...,H\}\times\{1,...,W\}$.  
In standard convolution, information is only exchanged among positions in a small neighborhood defined by the filter size (\eg, typically $3\times 3$). 
In order to create a large receptive field and capture long range dependencies, one needs to stack numerous layers after each other, as done in common architectures \cite{vgg,resnet}.
Graph convolution~\cite{gcnpaper}, is a highly efficient, effective and differentiable module that generalises the neighborhood definition used in standard convolution, and allows long-range information exchange in a single layer.
This is done by defining edges $\mathcal{E}$ among nodes $\mathcal{V}$ in a graph $\mathcal{G}$. 
Formally, following~\cite{gcnpaper}, graph convolution is defined as
\begin{equation}\label{eq:gcn}
		\tilde{\mathbf{X}} =  \sigma(\mathbf{A} \mathbf{X} \mathbf{W}),
\end{equation}
where $\sigma(\cdot)$ is the non-linear activation function, $\mathbf{A}\in \mathbb{R}^{N\times N}$ is the adjacency matrix characterising the neighbourhood relations of the graph and  $\mathbf{W}\in\mathbb{R}^{D \times \tilde{D}}$ is the weight matrix. 
Clearly, the graph definition and structure play a key role in determining the information propagation.
Our proposed framework is motivated by building orthogonal graph spaces via different graph projection strategies to learn a better task-specific representation.
As summarised in Fig.~\ref{fig:net}, we now describe how we propagate information in the coordinate space in Sec.~\ref{sec:method_coordinate_space}, and in feature space in Sec.~\ref{sec:method_feature_space}.

\subsection{Graph convolution in coordinate space}
\label{sec:method_coordinate_space}
\noindent{\textbf{Coordinate space projection.}} 
We first project the input feature into a new coordinate space $\Omega_\mathcal{S}$.
In general, we adopt a spatial downsampling operation $\mathbf{H}_\mathcal{S}$ to transform the input feature $\mathbf{X}$ to a new feature $\mathbf{V}_\mathcal{S} \in \mathbb{R}^{\frac{H \times W}{d^2} \times D}$ in the new coordinate space $\Omega_\mathcal{S}$, where the $d$ denotes the downsample rate,
\begin{equation}\label{eq:coordfeat}
		\mathbf{V}_\mathcal{S} = \mathbf{H}_\mathcal{S} \mathbf{X}.
\end{equation}

We consider two different operations for $\mathbf{H}_\mathcal{S}$: 
(1). Parameter-free operation. We take the downsampling operation $\mathbf{H}_\mathcal{S}$ to be average pooling which requires no additional learnable parameters.
(2). Parameterized operation. For efficiency, a downsampling rate of $d$ is achieved by chaining 
$\log_2{(d)}$ depth-wise convolution layers, each with a stride of 2 and kernel size of $3 \times 3$.

\noindent{\textbf{Coordinate graph convolution.}} 
After projecting the features into the new coordinate space $\Omega_\mathcal{S}$, we can build a lightweight fully-connected graph with adjacency matrix $\mathbf{A}_\mathcal{S} \in \mathbb{R}^{\frac{H}{d} \times \frac{W}{d}}$ for diffusing information across nodes.
Note that the nodes of the graph aggregate information from a ``cluster'' of pixels, and the edges measure the similarity between these clusters. 

The global relational reasoning is performed on the downsampled feature $\mathbf{V}_\mathcal{S}$ to model the interaction between the features of the corresponding nodes.
In particular, we adopt three learnable linear transformations ($\delta$, $\psi$, $\upsilon$) on the feature $\mathbf{V}_\mathcal{S}$ to produce the message, $\mathbf{M}_{\mathcal{S}}$
\begin{equation}\label{eq:message_spatial}
		{\mathbf{M}}_\mathcal{S} =  \mathit{f}(\delta(\mathbf{V}_\mathcal{S}) ,~\psi(\mathbf{V}_\mathcal{S})^\top) \upsilon(\mathbf{V}_\mathcal{S}) \mathbf{W}_\mathcal{S},
\end{equation}
where $\mathit{f}(\delta(\mathbf{V}_\mathcal{S}) ,~\psi(\mathbf{V}_\mathcal{S})^\top)$ gives the adjacency matrix in Eq.~\ref{eq:gcn},  $\mathit{f}$ is the dot-product operation.

\noindent{\textbf{Reprojection.}} 
After the reasoning, we map the new features $\mathbf{M}_\mathcal{S}$ back into the original coordinate space ($\mathbb{R}^{H \times W\times D}$) to be compatible with a regular CNN. 
Opposite to the downsample operation $\mathbf{H}_\mathcal{S}$ used in graph projection, we simply perform \emph{upsampling} as the reprojection operation. 
In particular,
nearest neighbour interpolation, $\mathit{interp}(\cdot)$, is adopted to resize $\mathbf{M}_\mathcal{S}$ to the original spatial input size ($N$). 
Hence, the output map is computed as $\tilde{\mathbf{X}}_\mathcal{S} = \xi(\mathit{interp}( \mathbf{M}_\mathcal{S}))$, where $\xi(\cdot)$ is a $1\times1$ convolution that transforms $\mathbf{M}_\mathcal{S}$ into the channel dimension $D$.

\noindent{\textbf{Discussion.}} 
Our coordinate GCN is built in a coarser spatial grid and its size is determined by the downsampling rate $d$ (we usually set $d=8$, the effect of changing the downsample rate is analysed in Section~\ref{sec:ablation}).
Compared to the Non-local operator~\cite{Nonlocal} that needs to build a large fully-connected graph with adjacency matrix $\mathbf{A}\in \mathbb{R}^{HW\times HW}$, our method is significantly more efficient.
Moreover, by re-ordering Eq.~\ref{eq:message_spatial} to ${\mathbf{M}}_\mathcal{S} =  \delta(\mathbf{V}_\mathcal{S}) (\psi(\mathbf{V}_\mathcal{S})^\top \upsilon(\mathbf{V}_\mathcal{S}))\mathbf{W}_\mathcal{S}$ (following the associative rule), we can obtain large savings in terms of memory and computation (from $O((HW)^2)$ to $O(HW)$).

\subsection{Graph convolution in feature space}
\label{sec:method_feature_space}
Given that the coordinate space GCN explicitly models the spatial relationships between pixels in the image, we now consider projecting the input feature into the feature space $\mathcal{F}$.
The coordinate space GCN enables our network to produce coherent predictions that consider all objects in the image, whilst the feature space GCN models interdependencies along the channel dimensions of the network's feature map.
Assuming that filters in later layers of a FCN are responsive to object parts and other high-level features \cite{zeiler2014visualizing}, then the feature space GCN captures correlations between more abstract features in the image like object parts.
\noindent{\textbf{Feature space projection.}} 
In practice, we first reduce the dimension of the input feature $\mathbf{X}$ with function $\theta(\mathbf{X}) \in \mathbb{R}^{N \times D_1}$ and formulate the projection function $\varphi(\mathbf{X}) = \mathbf{H}_\mathcal{F}^\top \in \mathbb{R}^{N \times D_2}$ as a linear combination of input $\mathbf{X}$ such that the new features can aggregate information from multiple regions.

Formally, the input feature $\mathbf{X}$ is projected to a new feature $\mathbf{V}_\mathcal{F}$ in the feature space $\mathcal{F}$ via the projection function $\mathbf{H}_\mathcal{F}^\top$. 
Thus we have
\begin{equation}\label{eq:feature}
		\mathbf{V}_\mathcal{F} = \mathbf{H}_\mathcal{F}^\top \theta(\mathbf{X}) = \varphi(\mathbf{X}) \theta(\mathbf{X}),
\end{equation}
where both functions of $\theta(\cdot)$ and $\varphi(\cdot)$ are implemented with $1 \times 1$ convolutional layer. 
This results in a new feature $\mathbf{V}_\mathcal{F} \in \mathbb{R}^{D_2 \times D_1}$, which consists of $D_2$ nodes, each of dimension $D_1$.

\noindent{\textbf{Feature graph convolution.}} 
After projection, we can build a fully-connected graph with adjacency matrix $\mathbf{A}_\mathcal{F} \in \mathbb{R}^{D_2 \times D_2}$ in the feature space $\mathcal{F}$, where each node contains the feature descriptor.
Following Eq.~\ref{eq:gcn}, we have
\begin{equation}\label{eq:fgcn}
		{\mathbf{M}}_\mathcal{F} =  (\mathbf{I} - \mathbf{A}_\mathcal{F})\mathbf{V}_\mathcal{F}\mathbf{W}_\mathcal{F},
\end{equation}
where $\mathbf{W}_\mathcal{F} \in \mathbb{R}^{D_1\times D_1}$ denotes the layer-specific trainable edge weights. 
We consider Laplacian smoothing~\cite{li2018deeper, graph_reason} 
by updating the adjacency matrix to $(\mathbf{I} - \mathbf{A}_\mathcal{F})$ to propagate the node features over the graph. 
The identity matrix $\mathbf{I}$ serves as a residual \emph{sum} connection in our implementation that alleviates optimisation difficulties. 
Both adjacency matrix $\mathbf{A}_\mathcal{F}$ and $\mathbf{W}_\mathcal{F}$ are randomly initialised and optimised by gradient descent during training in an end-to-end fashion.

\noindent{\textbf{Reprojection.}} 
As in Sec.~\ref{sec:method_coordinate_space}, after the reasoning, we map the new features $\mathbf{M}_\mathcal{F}$ back into the original coordinate space with output $\tilde{\mathbf{X}}_\mathcal{F} \in \mathbb{R}^{N \times D}$ to be compatible with regular convolutional neural networks,
\begin{equation}\label{eq:fgcnr}
		\tilde{\mathbf{X}}_\mathcal{F} = \phi(\mathbf{H}_\mathcal{F} \mathbf{M}_\mathcal{F}).
\end{equation} 
This is done by first reusing the projection matrix $\mathbf{H}_\mathcal{F}$ and then performing a linear projection (\eg, $1\times1$ convolution layer) to transform  $\tilde{\mathbf{M}}_\mathcal{F}$ into the original coordinate space. 
As a result, we have the feature $\tilde{\mathbf{X}}_\mathcal{F}$ with feature dimension of $D$ at each grid coordinate.

\subsection{DGCNet}
\label{sec:network_arch}

The final refined feature is computed as 
$\tilde{\mathbf{X}} =  \mathbf{X} + \tilde{\mathbf{X}}_\mathcal{S} + \tilde{\mathbf{X}}_\mathcal{F}$,
where ``+'' denotes point-wise summation.
To this end, we can easily incorporate our proposed module into existing backbone CNN architectures (\eg, ResNet-101).
Figure~\ref{fig:net} shows the schematic illustration of our proposed DGCNet .

\noindent{\textbf{Implementation of DGCNet.}} 
We insert our proposed module between two $3 \times 3$ convolution layers (both layers output $D = 512$ channels) appended at the end of a Fully Convolutional Network (FCN) for the task of semantic segmentation.
Specifically, we use an ImageNet pretrained ResNet-101 as backbone network, removing the last pooling and FC layer.
Our proposed module is then random initialised.
Dilated convolution and multi-grid strategies~\cite{DAnet} are adopted in last two stages of the backbone.
We simply set $D_1 = \frac{D}{2}$ and $D_2 = \frac{D}{4}$ in our implementation.
We add a synchronised batch normalisation (BN) layer and ReLU non-linearity after each convolution layer in our proposed module, except for the convolution layers in the coordinate space GCN (\ie, ~there are no BN and ReLU operations in the coordinate space GCN defined in Sec.~\ref{sec:method_coordinate_space}). 
\section{Experiments}
\label{sec:exp}
To evaluate our proposed method, we carry out comprehensive experiments on the Cityscapes \cite{Cityscapes} and PASCAL Context \cite{mottaghi_cvpr14} datasets, where we achieve state-of-art performance.
We describe our experimental setup in Sec.~\ref{sec:exp_setup}, before presenting experiments on the Cityscapes dataset (on which we also perform ablation studies) in Sec.~\ref{sec:exp_cs} and finally Pascal Context in Sec.~\ref{sec:exp_pc}.

\subsection{Experimental setup}
\label{sec:exp_setup}

\textbf{Cityscapes:} 
This dataset \cite{Cityscapes} densely annotates 19 object categories in urban scenes captured from cameras mounted on a car.
It contains 5000 finely annotated images, split into 2975, 500 and 1525 for training, validation and testing respectively.
The images are all captured at a high resolution of $2048 \times 1024$.

\noindent{\textbf{PASCAL Context:}} 
This dataset \cite{mottaghi_cvpr14} provides detailed semantic labels for whole scenes (both ``thing'' and ``stuff'' classes \cite{forsyth1996finding}), and contains 4998 images for training and 5105 images for validation.
Following previous works which we compare to, we evaluate on the most frequent 59 classes with along with one background category (60 classes) \cite{mottaghi_cvpr14}.

\noindent{\textbf{Implementation details:}}  We implement our method using Pytorch. Following \cite{pspnet}, we use momentum and adopt a polynomial learning rate decay schedule where the initial learning rate is multiplied by $(1 - \frac{\text{iter}}{\text{total}\_\text{iter}})^{0.9}$. 
The initial learning rate is set to 0.01 for Cityscapes and 0.001 for Pascal Context. 
Momentum and weight decay coefficients are set to 0.9 and 0.0001 respectively. 
For data augmentation, we apply random cropping (crop size with 769) and random left-right flipping during training for Cityscapes. 
For the Pascal Context dataset, our crop size is 480. 
We also use synchronised batch normalisation for better estimation of the batch statistics.

\noindent{\textbf{Metrics:}} Following the common procedure of \cite{VOC,ADE20K,Cityscapes}, we report the mean Intersection over Union (IoU), averaged over all classes.

\begin{table}
 \centering
  \caption{Ablation studies on (a) the proposed components of our network and (b) additional training and inference strategies. All methods use the ResNet-101 backbone, and are evaluated using the mean IoU on the Cityscapes validation set. Refer to Sec.~\ref{sec:ablation} for additional details.}
  \vspace{0.2cm}
 \begin{minipage}{\dimexpr.54\linewidth}
  \centering
  \small
  \resizebox{1.0\textwidth}{!}{%
   \begin{tabular}{l|p{1.6cm}<{\centering}|p{1.7cm}<{\centering}|p{1.7cm}<{\centering}|p{1.4cm}<{\centering}}
\toprule
& Backbone & Coord. GCN & Feature GCN & mIoU (\%)   \\
\midrule
Dilated FCN   &ResNet-101  &\xmark &\xmark & 75.2 \\
GCN & ResNet-101  & \cmark &\xmark  &78.8   \\
GCN &ResNet-101   &\xmark  & \cmark  &79.3   \\
DGCNet &ResNet-101   &\cmark & \cmark   &80.5  \\
\bottomrule
\end{tabular}
}
  \par
  {\footnotesize(a) Comparison of different graph modules.}
 \end{minipage}
 \begin{minipage}{\dimexpr.43\linewidth}
  \centering
  \small
  \resizebox{1.0\textwidth}{!}{%
   \begin{tabular}{l|p{1.2cm}<{\centering}|p{1.5cm}<{\centering}|p{0.8cm}<{\centering}|p{1.4cm}<{\centering}}
\toprule
 & OHEM & Multi-grid &MS &mIoU (\%)   \\
\midrule
DGCNet   &\xmark  &\xmark &\xmark & 79.5 \\
DGCNet & \cmark   & \xmark &\xmark  &79.8   \\
DGCNet &\cmark   &\cmark  & \xmark  &80.5   \\
DGCNet &\cmark  &\cmark & \cmark   &81.8   \\
\bottomrule
\end{tabular}
  }\par
  {\footnotesize(b) Additional training and inference strategies.}
 \end{minipage}
\label{tab:ablation_model}
\end{table}
\begin{table}
 \centering
  \caption{Ablation studies on (a) computational cost (input size [$1\times512\times128\times128$] and (b) graph projection strategy. All methods are evaluated using the mean IoU at a single scale, using the ResNet-101 backbone on the Cityscapes validation set. }
  \vspace{0.2cm}
 \begin{minipage}{\dimexpr.54\linewidth}
  \centering
  \small
  \resizebox{1.0\textwidth}{!}{%
   \begin{tabular}{l|p{1.1cm}<{\centering}|p{1.3cm}<{\centering}|p{1.1cm}<{\centering}}
\toprule
  & GFLOPs & \#Params &mIoU   \\
\midrule
DA module~\cite{DAnet} & 24.87 & 1 496 224 &  79.8 \\
DGC module (Ours) & 14.15 & 1 240 704 & 80.5 \\
\bottomrule
\end{tabular}
}
  \par
  {\footnotesize(a) Comparison on computation cost.}
 \end{minipage}
 \begin{minipage}{\dimexpr.45\linewidth}
  \centering
  \small
  \resizebox{1.0\textwidth}{!}{%
   \begin{tabular}{l|p{0.8cm}<{\centering}|p{0.8cm}<{\centering}|p{0.8cm}<{\centering}}
\toprule
  Downsample rate& d=4 & d=8 & d=16   \\
\midrule
Avg. pooling &80.2  & 80.5 & 80.5 \\
Stride conv. &80.0  & 80.5  &80.5   \\
\bottomrule
\end{tabular}
  }\par
  {\footnotesize(b) Ablation on graph projection strategies.}
 \end{minipage}
 \vspace{3mm}
 \vspace{-5mm}
 \label{tab:ablation_graph_projection}
\end{table}

\subsection{Experiments on Cityscapes}
\label{sec:exp_cs}

\subsubsection{Ablation Studies}
\label{sec:ablation}

\noindent{\bfseries Effect of proposed modules:} As shown in Table \ref{tab:ablation_model}a, our proposed GCN modules substantially improve performance.
Compared to our baseline FCN with dilated convolutions (with the ResNet-101 backbone), appending the Channel-GCN module obtains a mean IoU of 79.3\%, which is an improvement of 4.1\%.
Similarly, the Spatial-GCN module on its own improves the baseline by 3.6\%.
The best results are obtained by combining the two modules together, resulting in a mean IoU of 80.5\%.
The effect of our modules are visualised in Fig.~\ref{fig:city_res}.
The contextual information captured by our graph module improves the consistency of our results leading to fewer artifacts in the prediction.

\noindent{\bfseries Effect of additional training and inference strategies:} 
It is common to use additional ``tricks'' to improve results for semantic segmentation benchmarks \cite{deeplabv3,DAnet,ocnet,pspnet,ohem}.
Table \ref{tab:ablation_model}b shows that our proposed GCN module is complementary to them, and incorporating these strategies improves our results too.

Specifically, we consider 1) Online Hard Example Mining (OHEM) \cite{ohem,pohlen2017full,li2017holistic,ocnet} where the loss is only computed on the $K$ pixels with the highest loss in the image. Following \cite{ocnet}, we used 
$K = 10^{5}$ in a $769 \times 769$ cropped training image
2) Multi-Grid \cite{deeplabv3,DAnet,seg_vplr_zhu_cvpr2019} employs a hierarchy of convolutional filters of different dilation rates (4, 8 and 16) in the last ResNet block. 
3) Multi-Scale (MS) ensembling is commonly used at inference time to improve results \cite{deeplabv1,deeplabv3p,pspnet,ocnet,DAnet,psanet,boxsup}
We average the segmentation probability maps from 6 image scales\{0.75, 1 1.25, 1.5, 1.75, 2\} during inference.

As shown in Table.~\ref{tab:ablation_model}b each of these strategies provides consistent improvements to our overall results.
Using these strategies, we compare to the state-of-art in the following subsection.

\begin{figure*}
	\centering
	\includegraphics[width=0.95\linewidth]{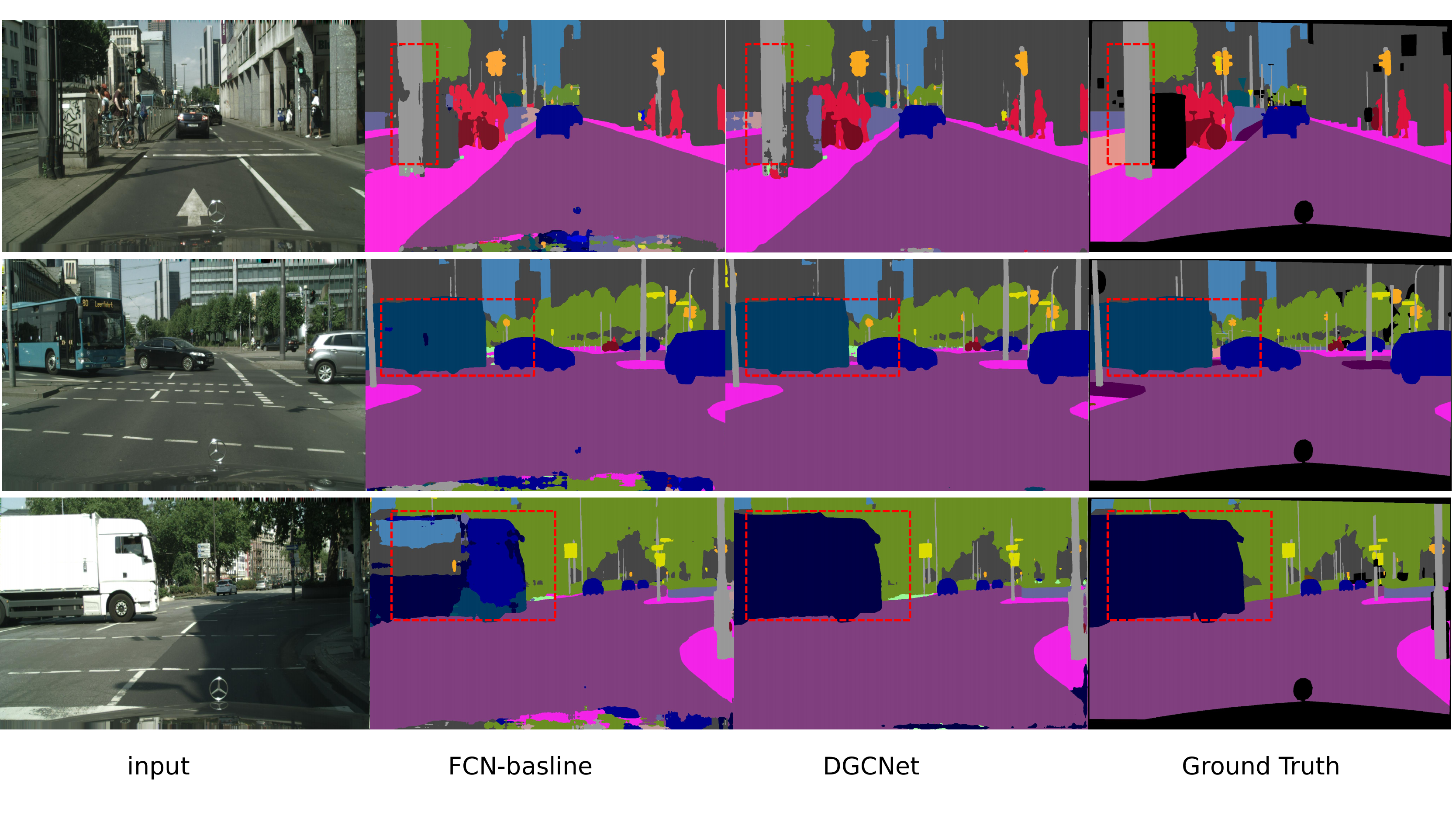}
	\vspace{-4mm}
	\caption{
		Cityscapes results compared with Dilated FCN ResNet101 baseline \cite{dilation}. Red boxes show our method can handle inconsistent results within the same object.
		Best view in color. 
		More results in supplementary material.}
	\label{fig:city_res}
\end{figure*}

\begin{table}[t]
    \begin{minipage}{\dimexpr.49\linewidth}
    \renewcommand\arraystretch{1.1}
    \small
    \caption{State-of-the-art comparison on the Cityscapes test set.}
    \begin{center}\scalebox{0.8}{
    \begin{tabular}{l|p{1.8cm}<{\centering}|p{1.3cm}<{\centering}}
    \toprule
    Method & Backbone & mIoU (\%)   \\
    \midrule
    PSPNet~\cite{pspnet}   &ResNet-101  &78.4\\
    PSANet~\cite{psanet} & ResNet-101  & 78.6  \\
    OCNet~\cite{ocnet} & ResNet-101 & 80.1 \\
    DGCNet (Ours) \textsuperscript{\textdagger} & ResNet-101   &\bf  80.9   \\
    \bottomrule
    SAC~\cite{sac} &ResNet-101   &78.1   \\
    AAF~\cite{aaf} &ResNet-101  &79.1  \\
    BiSeNet~\cite{bisenet} &ResNet-101   &78.9   \\
    PSANet~\cite{psanet} &ResNet-101  &80.1   \\
    DFN~\cite{dfn} &ResNet-101   &79.3  \\
    DepthSeg~\cite{depthseg} &ResNet-101  &78.2  \\
    DenseASPP~\cite{denseaspp} &ResNet-101   &80.6   \\
    GloRe~\cite{graph_reason} &ResNet-101  &80.9  \\
    DANet~\cite{DAnet} &ResNet-101     &81.5   \\
    OCNet~\cite{ocnet} & ResNet-101 & 81.7 \\
    DGCNet (Ours) \textsuperscript{\textdaggerdbl} & ResNet-50 & 80.8 \\
    DGCNet (Ours) \textsuperscript{\textdaggerdbl} &ResNet-101   &\bf 82.0   \\
    \bottomrule
    \end{tabular}}
    \end{center}
    \vspace{-2mm}
    \footnotesize{
    \textsuperscript{\textdagger}: trained only on train-fine set.
    \\
    \textsuperscript{\textdaggerdbl}: trained on train-fine and val-fine sets.
    }
    \label{tab:citytest}
    \end{minipage}
    \hspace{0.1cm}
    \begin{minipage}{\dimexpr.49\linewidth}
    \small
    \vspace{-\baselineskip}
    \caption{Comparison to other methods on Pascal Context \cite{mottaghi_cvpr14} dataset.}
    \begin{center}\scalebox{0.8}{
    \begin{tabular}{l|p{1.8cm}<{\centering}|p{1.3cm}<{\centering}}
    \toprule
    Method & Backbone & mIoU (\%)   \\
    \midrule
    FCN8-s~\cite{fcn} & VGG-16 & 37.8 \\
    HO CRF~\cite{arnab2016higher} & VGG-16 & 41.3 \\
    Piecewise~\cite{lin2016efficient} & VGG-16  & 43.3 \\
    DeepLab-v2 (COCO)~\cite{deeplabv2}& ResNet-101   &45.7   \\
    RefineNet~\cite{refinenet}&ResNet-101   &47.3  \\
    PSPNet~\cite{pspnet}&ResNet-101   &47.8  \\
    Ding~\textit{et al.}~\cite{ding2018context} &ResNet-101  &51.6  \\
    EncNet~\cite{encodingnet} (SS) &ResNet-50   &49.0  \\
    EncNet~\cite{encodingnet} (MS) &ResNet-101   &51.7  \\
    SGR~\cite{SGR_gcn} &ResNet-101   &52.5 \\
    DANet~\cite{DAnet}&ResNet-50  &50.1   \\
    DANet~\cite{DAnet}&ResNet-101  &52.6   \\
    \bottomrule
    Dilated FCN baseline &ResNet-50  &44.3 \\ 
    DGCNet (SS) &ResNet-50   &50.1  \\ 
    DGCNet (SS) &ResNet-101  &\bf 53.0 \\ 
    DGCNet (MS) &ResNet-101     &\bf 53.7\\ 
    \bottomrule
    \end{tabular}}
    \end{center}
    \footnotesize{
    SS: Single scale. MS: Multi scale
    }
    \label{tab:context}
    \end{minipage}
\end{table}

\begin{figure*}[t]
	\centering
	\includegraphics[width=0.95\linewidth]{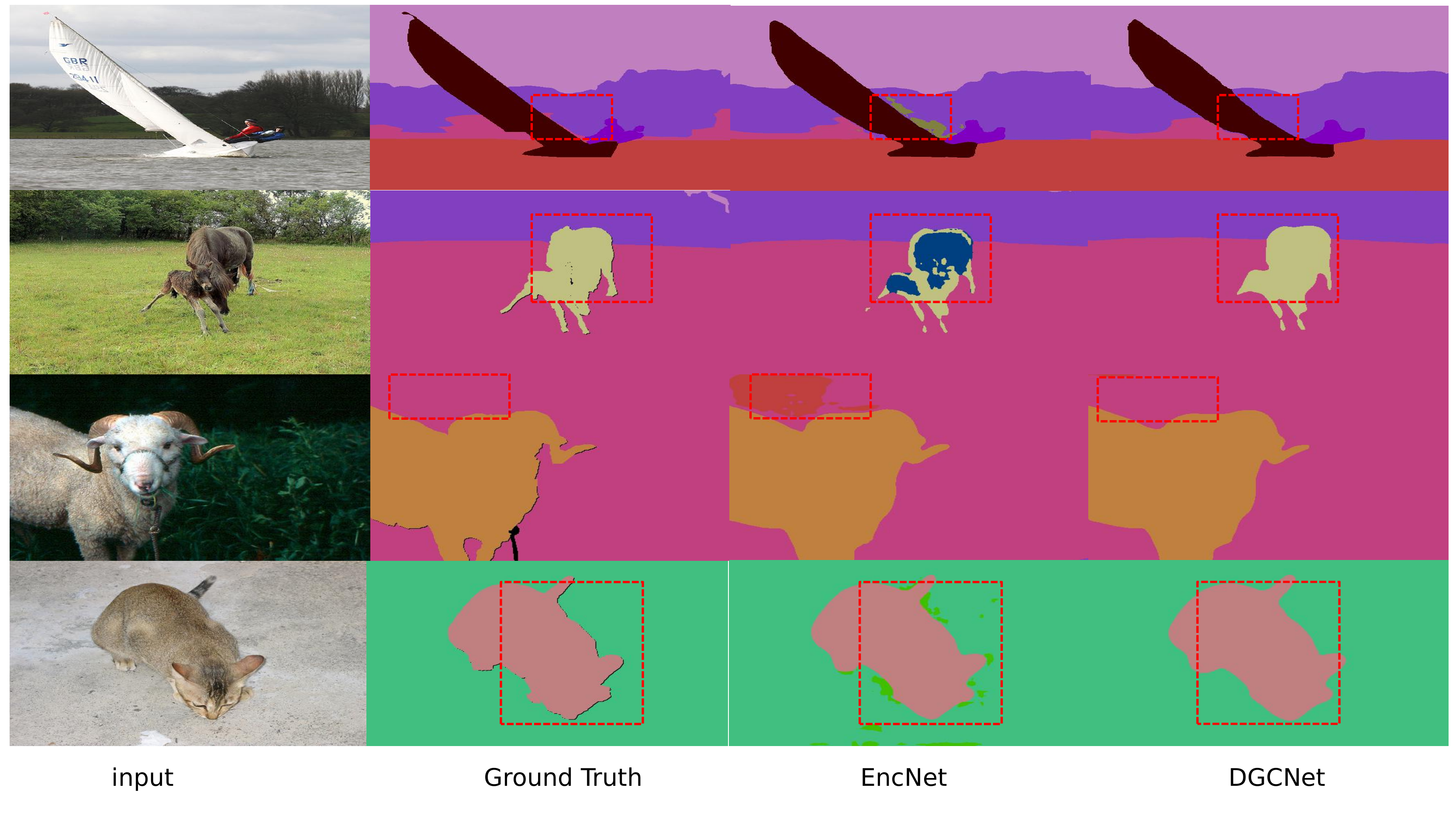}
	\caption{
	    Comparison of our results on Pascal Context to the state-of-art EncNet \cite{encodingnet} method.
	    Note that our results are more consistent and have fewer artifacts.
		Best view in color.
		More results in supplementary material.}
	\label{fig:pcontext_res}
\end{figure*}

\noindent{\bfseries Computational cost analysis} 
Table \ref{tab:ablation_graph_projection}a shows that our proposed method costs significantly fewer floating point operations than the related work of DANet \cite{DAnet}, but still achieves higher performance.

\noindent{\bfseries Effect of mapping strategies:} 
As mentioned in Sec.~\ref{sec:method_coordinate_space}, different mapping strategies are possible to build the coordinate space graph.
We present the effect of two strategies -- average pooling and strided convolution -- for different downsampling ratios, $d$, in Tab.~\ref{tab:ablation_graph_projection}b.
It is interesting to observe that the model with average pooling achieves similar performance as strided convolution, even slightly outperforming it for $d = 4$.
One reason may be that the model with the parameter-free mapping strategy can overfit less than the one using the strided convolution operation.
Note that both strategies are robust to the choice of $d$, and similar performance is obtained for $d = 4$, $8$ and $16$.
The final model that we use for comparing to the state-of-the-art uses strided convolutions with $d = 8$.

\subsubsection{Comparisons with state-of-the-art}
Table \ref{tab:citytest} compares our approach with existing methods on the Cityscapes test set.
Following common practice towards obtaining the highest performance, we use the inference strategies described in the previous section.
For fair comparison, Tab.~\ref{tab:citytest} shows methods that are only trained using the fine annotations from Cityscapes, and evaluated on the evaluation server.
We achieve a mean IoU of 80.7\% when only using the training set, thus outperforming PSANet~\cite{psanet} by 2.1\% and OC-Net \cite{ocnet} by 0.8\%.
Training with both train-fine and val-fine sets achieves an IoU of 82.0\%.
In both scenarios, we obtain state-of-the-art results.
Detailed per-class results are provided in the supplementary material, which shows that our method achieves the highest IoU in 16 out of the 19 classes.

\subsection{Experiments on Pascal Context}
\label{sec:exp_pc}

Table~\ref{tab:context} shows our results on Pascal Context.
We follow prior work~\cite{refinenet,encodingnet,DAnet} to use the semantic labels of the most frequent 59 object categories plus background (therefore, there are 60 classes in total).
The Dilated-FCN baseline achieves a mean IoU of 44.3\% with the ResNet-50 backbone.
Our proposed DGCNet significantly improves this baseline, achieving an IoU of 50.1\% with the same ResNet-50 backbone under single scale evaluation (SS), which outperforms previous work using the same backbone (49.0) \cite{encodingnet}.
With the ResNet-101 backbone, DGCNet achieves an IoU of 53.0\%. 
Moreover, our performance further improves to 53.7\% when multiscale inference (MS) is adopted, surpassing the previous state-of-the-art~\cite{DAnet} by a large margin.

\begin{table}
\renewcommand\arraystretch{1.1}
\end{table}

\section{Conclusion}

We proposed a graph-convolutional module to model the contextual relationships in an image, which is critical for dense prediction tasks such as semantic segmentation.
Our method consists of two branches, one to capture context along the spatial dimensions, and another along the channel dimensions, in a convolutional feature map.
Our proposed approach provides significant improvements over a strong baseline, and achieves state-of-art results on the Cityscapes and Pascal Context datasets.
Future work is to address other dense prediction tasks such as instance segmentation and depth estimation.
\section*{Acknowledgments}
This work was supported by EPSRC Programme Grant Seebibyte EP/M013774/1, ERC grant ERC-2012-AdG 321162-HELIOS and EPSRC/MURI grant EP/N019474/1.
We gratefully acknowledge the use of the University of Oxford Advanced Research Computing (ARC) facility in carrying out this work.
We would also like to acknowledge the Royal Academy of Engineering, FiveAI and the support of DeepMotion AI Research for providing the computing resources in carrying out this research.
    
\clearpage
\appendix

\section*{Appendix}
\label{sec:app}

\vspace{0.8cm}
\begin{sidewaystable}[htbp]
	\renewcommand\arraystretch{1.3}
	\footnotesize
	\footnotesize
	\setlength{\tabcolsep}{2.0pt}
	\begin{center}
		\begin{tabular}{ l | c |c c c c c c c c c c c c c c c c c c c  c}
			\hline
			Methods &  \rotatebox{90}{Mean IoU} &  \rotatebox{90}{road} &  \rotatebox{90}{sidewalk} &  \rotatebox{90}{building} & \rotatebox{90}{ wall} &  \rotatebox{90}{fence} &  \rotatebox{90}{pole} & \rotatebox{90}{traffic light} &  \rotatebox{90}{traffic sign}&  \rotatebox{90}{vegetation} &  \rotatebox{90}{terrain} &  \rotatebox{90}{sky} & \rotatebox{90}{person} &  \rotatebox{90}{rider} & \rotatebox{90}{car} &  \rotatebox{90}{truck}& \rotatebox{90}{ bus}& \rotatebox{90}{ train}& \rotatebox{90}{ motorcycle}&  \rotatebox{90}{bicycle}\\
			\hline
			\hline
			DeepLab-v2~\cite{deeplabv2} & 70.4 & 97.9 & 81.3 & 90.3 & 48.8 & 47.4 & 49.6 & 57.9 & 67.3 & 91.9 & 69.4 & 94.2 & 79.8 & 59.8 & 93.7 & 56.5 & 67.5 & 57.5 & 57.7 & 68.8 \\
			RefineNet~\cite{refinenet} & 73.6 & 98.2 & 83.3 & 91.3 & 47.8 & 50.4 & 56.1 & 66.9 & 71.3 & 92.3 & 70.3 & 94.8 & 80.9 & 63.3 & 94.5 & 64.6 & 76.1 & 64.3 & 62.2 & 70 \\
			GCN~\cite{peng2017large} & 76.9 & - & - & - & - & - & - & - & - & - & - & - & - & - & - & - & - & - & - & -  \\ 
			DUC~\cite{wang2018understanding} & 77.6 & 98.5 & 85.5 & 92.8 & 58.6 & 55.5 & 65 & 73.5 & 77.9 & 93.3 & 72 & 95.2 & 84.8 & 68.5 & 95.4 & 70.9 & 78.8 & 68.7 & 65.9 & 73.8 \\  
			ResNet-38~\cite{wu2019wider} & 78.4 & 98.5 & 85.7 & 93.1 & 55.5 & 59.1 & 67.1 & 74.8 & 78.7 & 93.7 & 72.6 & 95.5 & 86.6 & 69.2 & 95.7 & 64.5 & 78.8 & 74.1 & 69 & 76.7 \\
			PSPNet~\cite{pspnet} & 78.4 & - & - & - & - & - & - & - & - & - & - & - & - & - & - & - & - & - & - & -  \\
			BiSeNet~\cite{yu2018bisenet} & 78.9 & - & - & - & - & - & - & - & - & - & - & - & - & - & - & - & - & - & - & -  \\ 
			PSANet~\cite{psanet} & 80.1 & - & - & - & - & - & - & - & - & - & - & - & - & - & - & - & - & - & - & -  \\ 
			DenseASPP~\cite{denseaspp} & 80.6 & \textbf{98.7} & 87.1 & 93.4 & 60.7 & 62.7 & 65.6 & 74.6 & 78.5 & 93.6 & 72.5 & 95.4 & 86.2 & 71.9 & 96.0 & \textbf{78.0} & 90.3 & 80.7 & 69.7 & 76.8\\    
			GloRe~\cite{graph_reason} & 80.9 & - & - & - & - & - & - & - & - & - & - & - & - & - & - & - & - & - & - & -  \\   
			DANet~\cite{DAnet} & 81.5 & 98.6 & 86.1 & \textbf{93.5} & 56.1 & 63.3 & 69.7 & 77.3 & \textbf{81.3} & 93.9 & 72.9 & 95.7 & 87.3 & 72.9 & 96.2 & 76.8 & 89.4 & \textbf{86.5} & \textbf{72.2} & \textbf{78.2}\\      
			\hline 
			Ours & \textbf{82.0} & \textbf{98.7} & \textbf{87.4} & \textbf{93.9} & \textbf{62.4} & \textbf{63.4} & \textbf{70.8} & \textbf{78.7} & \textbf{81.3} & \textbf{94.0} & \textbf{73.3} & \textbf{95.8} & \textbf{87.8} & \textbf{73.7} & \textbf{96.4} & 76.0 & \textbf{91.6} & 81.6 & 71.5 & \textbf{78.2}\\     
			\hline
		\end{tabular}
	\end{center}
	\caption{Per-class results on Cityscapes testing set. Our methods outperforms existing approaches and achieves 82.0 \% in Mean IoU and achieves the highest IoU in 16 out of the 19 classes. }
	\vspace{-1.5em}
	\label{tab:citytest}
	\end{sidewaystable}

\begin{figure*}[ht]
	\centering
	\includegraphics[width=0.95\linewidth]{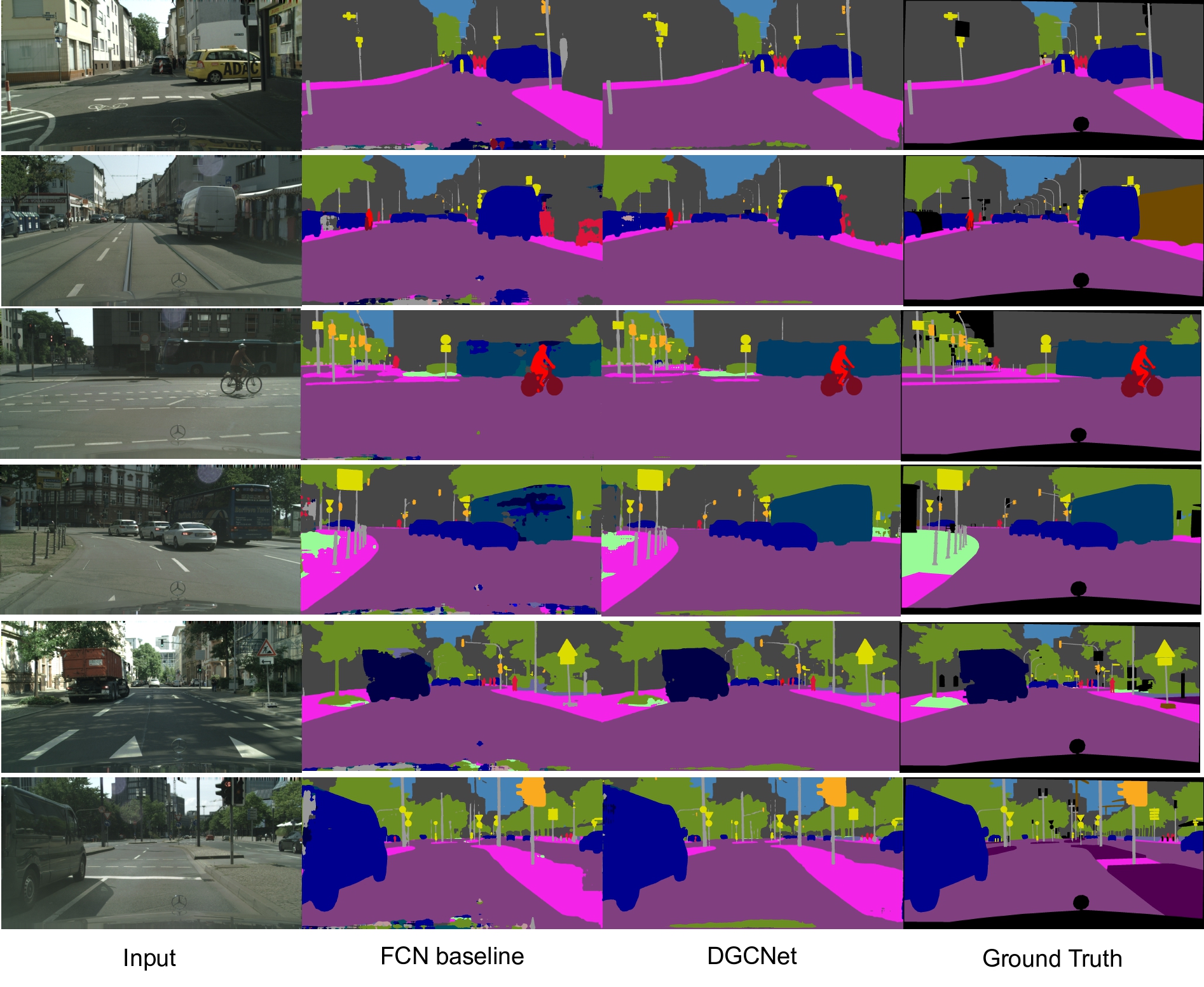}
	\vspace{-4mm}
	\caption{
		Cityscapes results compared with Dilated FCN ResNet101 baseline \cite{dilation}.
		Best view in color. }
	\label{fig:city_res}
\end{figure*}

\begin{figure*}[ht]
	\centering
	\includegraphics[width=0.95\linewidth]{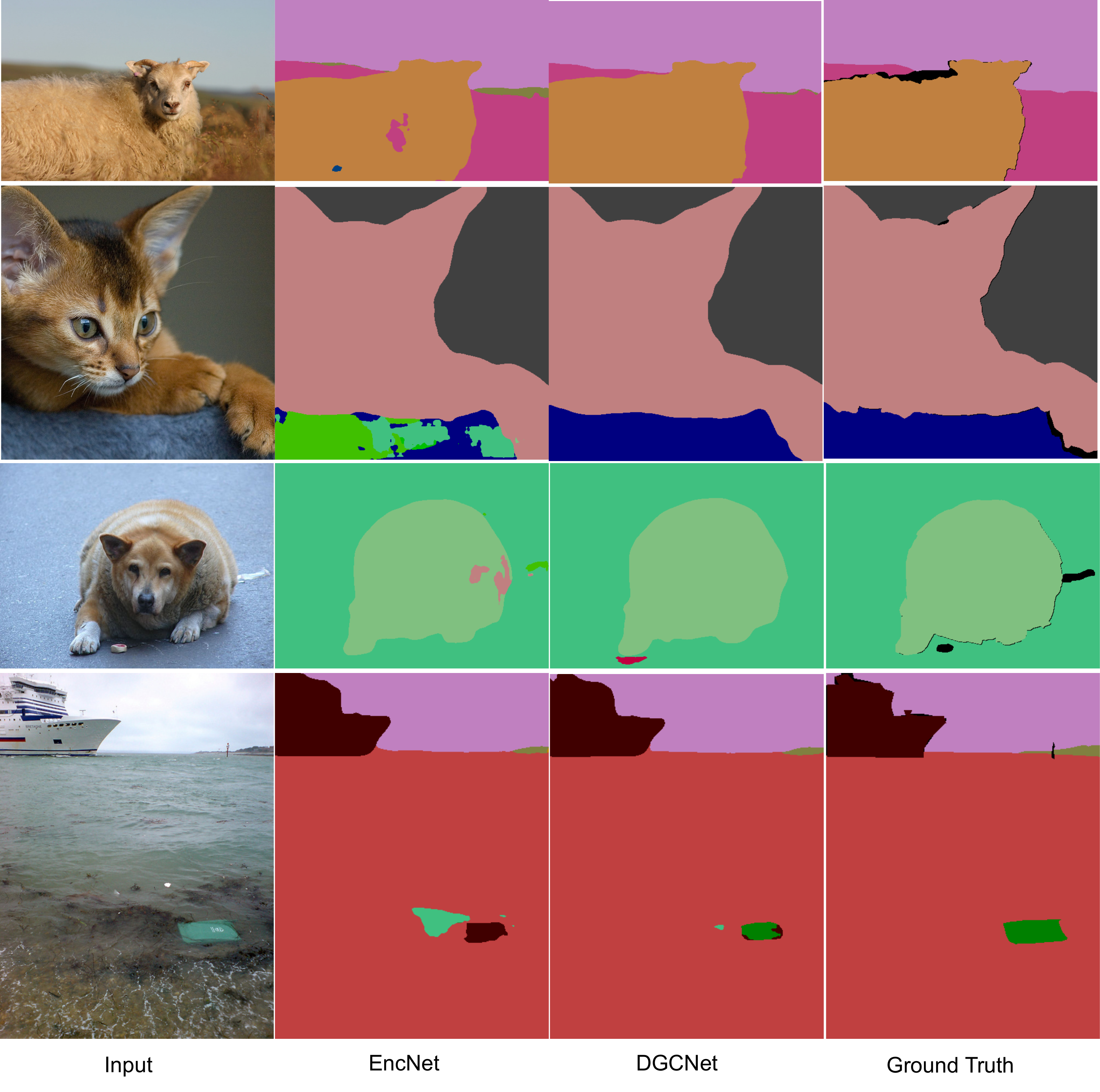}
	\vspace{-4mm}
	\caption{
		Comparison of our results on Pascal Context to the state-of-art EncNet \cite{encodingnet} method.
	    Note how our results are more consistent and have fewer artifacts.
		Best view in color. }
	\label{fig:city_res}
\end{figure*}

\clearpage
\bibliography{egbib}
\end{document}